\documentclass[letterpaper, 10 pt, conference]{ieeeconf}  % Comment this line out if you need a4paper

\IEEEoverridecommandlockouts         

\usepackage{graphicx}
\usepackage{amsmath, amssymb}
\usepackage{cite}
\usepackage{booktabs}
\usepackage{multirow}
\usepackage[table,xcdraw]{xcolor}
\usepackage{colortbl}
\usepackage{gensymb}
\usepackage{cuted}
\usepackage{caption}
\usepackage{hhline}

\title{How to Train your Tactile Model:\\ Tactile Perception with Multi-fingered Robot Hands}
\author{Christopher J. Ford, Kaichen Shi, Laura Butcher, Nathan F. Lepora, Efi Psomopoulou% <-this % stops a space
\thanks{CJF, LB, NFL and EP in Horizon Europe Project MANiBOT were supported by UKRI grant number 10080095. KS, NFL and EP were also supported by Royal Society International Collaboration Award ICA\textbackslash R1\textbackslash 241213 (South Korea). EP was also supported by ARIA Robot Dexterity award SMRB-PR01-P04/Artimus. \newline
\indent Authors are with the School of Engineering Mathematics and Technology and Bristol Robotics Laboratory, University of Bristol, UK (e-mail: [chris.ford, vr24150, dj23607, n.lepora, efi.psomopoulou]@bristol.ac.uk).}}%

\begin{document}

\maketitle

\begin{strip}\vspace{-5em}

        \centering
        \includegraphics[width = 0.99\textwidth]{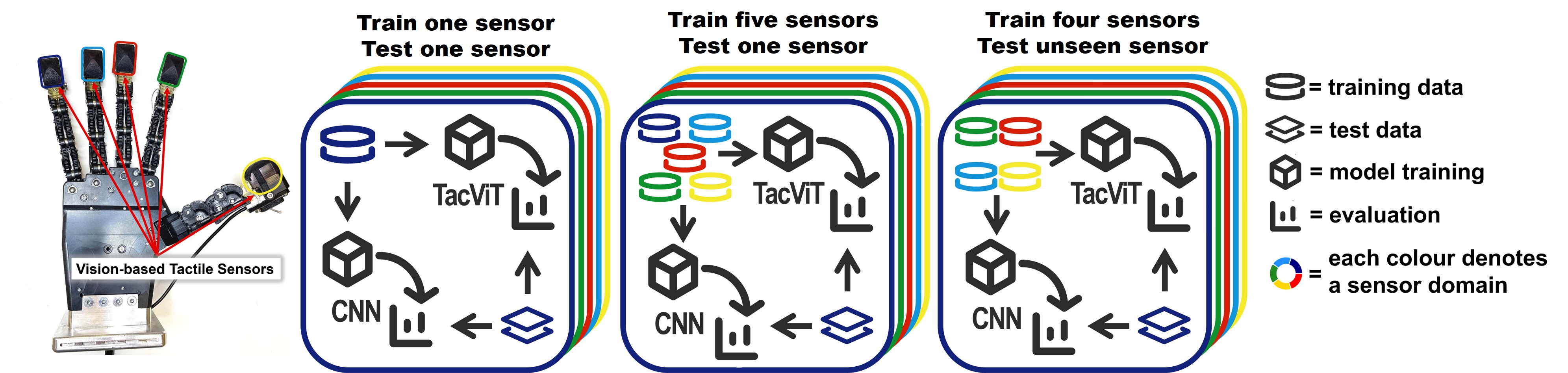}
        \captionof{figure}{Sensors from a five-fingered robot hand equipped with vision-based tactile sensors (VBTS) are used to evaluate TacViT and CNN models. Three experiments are conducted to assess model performance: training and testing on the same sensor (Tr1–Te1), training on all sensors and testing on one seen during training (Tr5–Te1), and training on four sensors and testing on an unseen sensor (Tr4–TeU). These experiments reveal that TacViT generalizes to new sensors without retraining.}
        \label{fig:demo}
        % \vspace{-1em}
\end{strip}

\begin{abstract}
Rapid deployment of new tactile sensors is essential for scalable robotic manipulation, especially in multi-fingered hands equipped with vision-based tactile sensors. However, current methods for inferring contact properties rely heavily on convolutional neural networks (CNNs), which, while effective on known sensors, require large, sensor-specific datasets. Furthermore, they require retraining for each new sensor due to differences in lens properties, illumination, and sensor wear. Here we introduce TacViT, a novel tactile perception model based on Vision Transformers, designed to generalize on new sensor data. TacViT leverages global self-attention mechanisms to extract robust features from tactile images, enabling accurate contact property inference even on previously unseen sensors. This capability significantly reduces the need for data collection and retraining, accelerating the deployment of new sensors. We evaluate TacViT on sensors for a five-fingered robot hand and demonstrate its superior generalization performance compared to CNNs. Our results highlight TacViT’s potential to make tactile sensing more scalable and practical for real-world robotic applications.

% \textcolor{red}{When working with multi-fingered robot hands with vision-based tactile sensors (VBTS), being able to quickly deploy a new fingertip sensor can be critical}. Traditional approaches \textcolor{red}{to infer contact properties}  predominantly rely on convolutional neural networks (CNNs), which, while effective on known sensors, struggle with \textcolor{red}{new sensor data} due to localized variations between different sensors of the same design. These variations, caused by differences in lens properties, illumination, and sensor wear, significantly impact model generalisation. In this paper, we introduce a novel method (TacViT) for \textcolor{red}{generalising on new sensor data} by leveraging a Vision Transformer (ViT) as a feature extractor. ViTs utilise a self-attention mechanism to capture global dependencies in tactile images, allowing for improved generalization across new sensors. We present the proposed architecture and evaluate its \textcolor{red}{generalising} performance against a CNN \textcolor{red}{for a five-fingered robot hand}, demonstrating TacViT's superior ability to generalize to unseen sensor data. Finally, we analyze the strengths and limitations of TacViT and propose potential avenues for future research to further enhance its adaptability and performance.

\end{abstract}

\section{Introduction}

The use of artificial tactile sensing is crucial in achieving intelligent robotic manipulation \cite{tegin2005tactile}, gaining increasing attention in recent years as contact-rich information cannot be acquired by solely vision sensing due to occlusions or varying lighting conditions \cite{giudici2024leveraging}. Vision-based tactile sensors (VBTS), like the TacTip \cite{ward2018tactip}, provide high-resolution images of contact interactions, enabling the use of vision-based methods to estimate contact pose, which may then be used for tactile-driven robot control \cite{lloyd2024pose}. TacTip sensors are low-cost and easily customisable as they are 3D printed, but their manufacturing process can introduce sensor-specific variations such as changes in lens properties, illumination conditions, and sensor wear over time.

\begin{figure}[h]
        \centering
        \includegraphics[width = 0.2\textwidth,angle=90,origin=c]{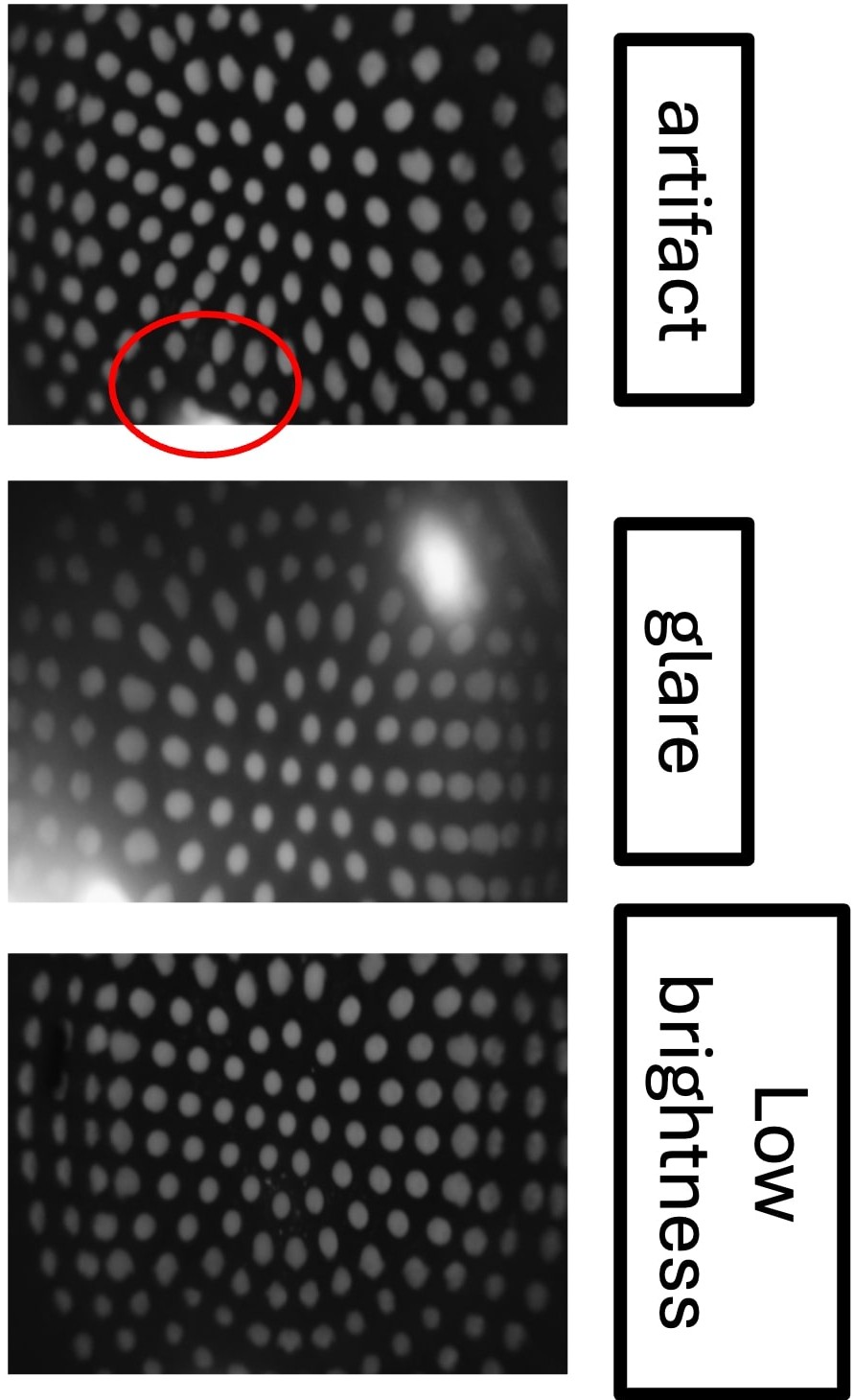}
        \vspace{-3em}
        \caption{Examples of tactile images from different sensors: Evident variation in the tactile images collected from different TacTip-style sensors.}
        \label{fig:tactip}
        \vspace{-1em}
\end{figure}

Current state-of-the-art methods for estimating contact pose from TacTip use convolutional neural networks (CNNs) to extract features from tactile images and use supervised deep learning to train a model to accurately predict contact variables \cite{lepora2020optimal}. While CNNs have demonstrated strong performance on known sensors (trained on data from a single sensor), their ability to transfer knowledge to unseen sensor data is limited, even if the new sensor is of the same form factor. This is due to CNN's reliance on local spatial hierarchies, which does not generalize well to sensor variations, even if these variations are minor. This is quite limiting when working on multi-fingered robot hands with tactile fingertips, where adding a new sensor would require collection of new data and training of a new model. In contrast, Vision Transformers (ViTs) \cite{dosovitskiy2020image} use self-attention mechanisms to model long-range dependencies, making them well-suited for generalising across multiple sensors. Unlike CNNs, which extract features in a hierarchical manner, ViTs process entire input images as sequences of patches, enabling more flexible feature learning. Recent work has demonstrated the potential of ViTs in various perception tasks \cite{patro2023spectformer, han2022survey}, however, their application in tactile sensing remains largely unexplored. Given the significant variants in different tactile images from TacTips of the same design (as shown in Fig. \ref{fig:tactip}), an approach for estimating contact variables which is robust to new unseen sensor data would be highly beneficial.

To address the challenge of deploying new tactile sensors without retraining, this paper introduces TacViT, a Vision Transformer-based model designed for tactile perception. TacViT uses a pre-trained ViT model fine-tuned on tactile images and labels for prediction of contact pose and force. Its performance is evaluated by testing it on sensors from a five-fingered robot hand, including an unseen new sensor to assess its generalisation capability (Fig. \ref{fig:demo}). Our contributions are:

\begin{enumerate}
    \item We propose TacViT, a ViT-based tactile perception model that generalizes to new, unseen sensors, maintaining accuracy and consistency.
    \item By reducing the need for sensor-specific data collection and retraining, TacViT enables faster and more scalable deployment of new tactile sensors in multi-fingered robotic hands.
    \item We validate TacViT’s performance through a series of experiments on tactile sensors from a five-fingered robot hand, comparing its accuracy and robustness against CNNs across multiple sensor configurations.
\end{enumerate}

\section{Background and Related Work}
Tactile sensing in robotics has evolved from single-point force sensors capable of providing basic contact information \cite{bicchi1989augmentation} to high-resolution, vision-based sensors \cite{lepora2021soft, abad2020visuotactile}. This transition has enabled access to richer contact information, but has also introduced new challenges related to processing this data to extract relevant information accurately and efficiently.

The Bristol Robotics Laboratory (BRL) TacTip (shown in Fig. \ref{fig:tactip}) is an artificial visuotactile sensor \cite{ward2018tactip}. The sensor is biomimetic to human touch, featuring an opaque rubber skin, on the underside of which is an array of contrasting markers cantilevered on pins. These pins are analogous to the subdermal papillae central to human tactile sensing \cite{lepora2021soft, dahiya2009tactile}. The volume beneath the skin is divided by an acrylic window and encapsulated by clear silicone gel. A camera is then used to observe the marker array, which shifts under deformation of the skin. The resulting tactile images may then be fed through various vision algorithms to extract contact information \cite{ward2018tactip}.

Earlier works with the TacTip focused on statistical and feature based methods for tactile perception \cite{james2018slip}. As GPU-accelerated methods became more widely available, the focus shifted to use of CNNs and supervised deep learning \cite{lloyd2024pose}. Whilst CNN-based approaches are able to output highly accurate contact pose predictions, they require thousands of domain-specific training data points for each new sensor configuration\cite{lloyd2024pose}. This is because minor differences in camera angles, lighting conditions, and sensor materials can introduce distribution shifts that significantly degrade CNN performance due to their focus on learning from localized features. This introduces a bottleneck, making large-scale deployment impractical.  \cite{rodriguez2023overcoming, lloyd2024pose}.

On the other hand, Vision Transformers (ViTs) offer a promising solution by learning more generalizable feature representations, reducing the dependency on domain-specific training data \cite{dosovitskiy2020image}. Whilst they do require large amounts of data for pre-training, they are able to be fine tuned on a much smaller, task specific dataset. They were originally developed for image classification tasks, where they demonstrated superior performance compared to CNNs on large-scale datasets such as ImageNet. By leveraging self-attention mechanisms, ViTs can model long-range dependencies without the constraints of local receptive fields. This enables them to generalize better across domains, as they are not biased towards specific spatial hierarchies.

Following their success in classification, ViTs have been extended to various regression-based tasks, including depth estimation \cite{zhao2022monovit}, global object pose estimation \cite{thalhammer2023self}, and robotic perception \cite{bonatti2023pact}. These applications demonstrate the flexibility of ViTs in tasks beyond classification, where precise spatial relationships between input features are crucial. Our work builds on this foundation by applying ViTs to VBTS, specifically for predicting depth and orientation from tactile images.

Our ViT for tactile perception (TacViT) builds upon recent advances in ViTs to improve tactile perception across multiple sensor domains. By leveraging self-attention mechanisms, TacViT captures global dependencies within tactile images, enabling it to generalize well to sensor domains unseen during training. This eliminates the need for extensive retraining when deploying new sensors, making tactile sensing more scalable for real-world applications. Our approach is particularly relevant as robotic manipulation systems move towards multi-fingered hands with distributed tactile sensing \cite{yang2024anyrotate}. Instead of training separate models for each sensor, TacViT provides a unified framework that can adapt to new sensor domains with minimal additional training.

\section{Methodology}

\begin{figure*}[t!]
        \centering
        \includegraphics[width = 0.75\textwidth]{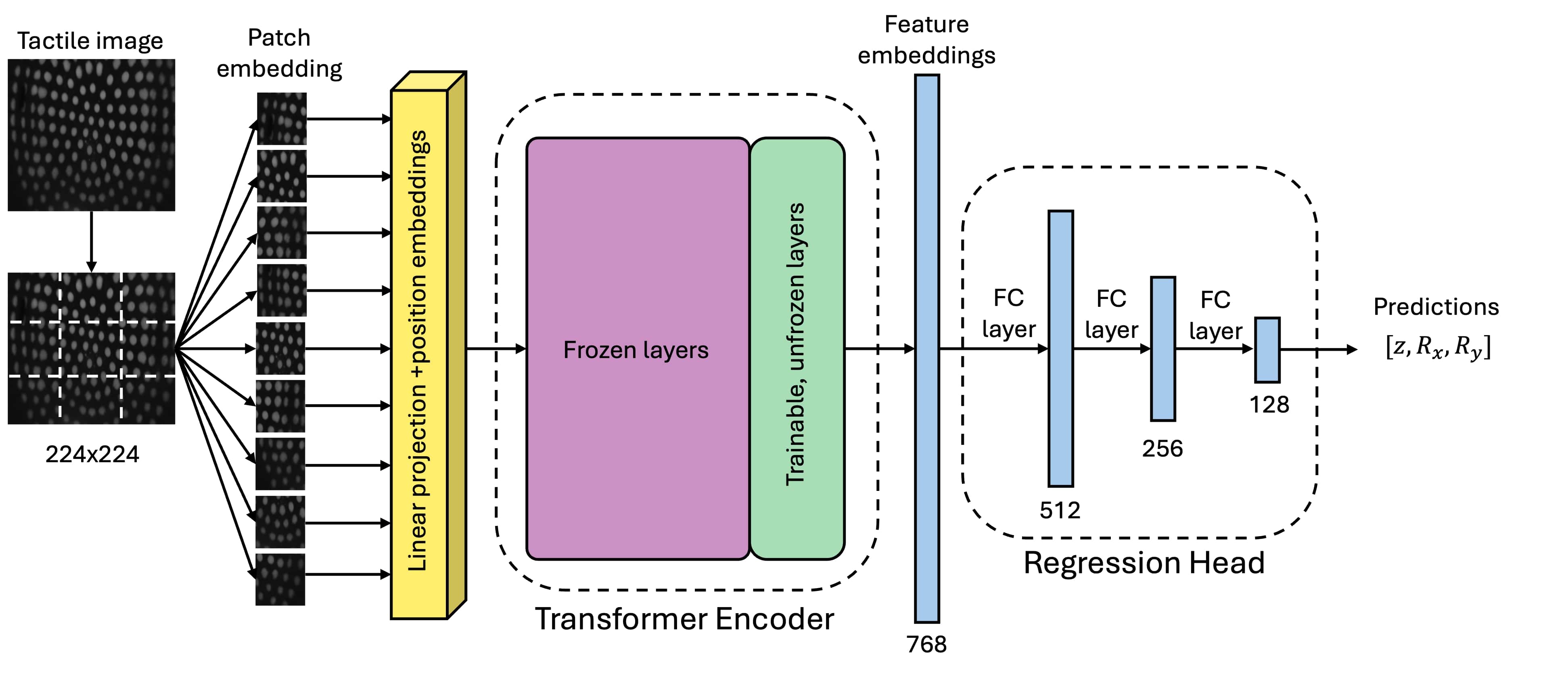}
        \caption{TacViT pipeline: a tactile image is divided into patches which are linearized and fed through a transformer encoder. The transformer extracts feature embeddings from the image, which is then passed through a regression head consisting of several fully connected (FC) layers to output pose predictions.}
        \label{fig:pipeline}
        \vspace{-1em}
\end{figure*}

\subsection{Model Architecture}
The architecture of TacViT (shown in Fig. \ref{fig:pipeline}) is divided into two distinct stages. First, a ViT model pre-trained on images from ImageNet (Google's vit-base-patch16-224) extracts feature embeddings from tactile images taken from multiple sensors in batches. These feature embeddings are then fed into a regression head along with pose and force labels corresponding to each sample in the batch, outputting predicted pose and force values. Training loss is then calculated by comparing the predicted values with the ground truth, which is backpropagated through the network to tune the weights. Details of the experiments, data collection process and training pipeline are provided in the following subsections.

\subsection{Evaluation Experiments}
To evaluate TacViT's ability to generalize to new sensors, we design experiments that simulate real-world deployment scenarios where a new sensor is introduced without prior training data. Three experiments were performed across multiple (five) fingertip tactile sensors using two model architectures, the TacViT model and a standard CNN model for comparison:

\subsubsection{Tr1-Te1: Model Performance on Single Sensors (baseline)}

To establish baseline performance, we first evaluate the architectures on data from a single sensor for five different fingertip sensors. This is done by training a model on tactile images from one sensor and testing it on tactile images from the same sensor (Tr1-Te1).

\subsubsection{Tr5-Te1: Model Performance on Multiple Known Sensors}
Once baseline performance is established, we evaluate the two model architectures on data across all five different fingertip sensors. This is done by training a model on tactile images from all five sensors and testing it on tactile images from one of the sensors seen in training (Tr5-Te1).

\subsubsection{Tr4-TeU: Model Performance on Unseen Sensors}
In the last experiment, we want to evaluate the ability of each architecture to generalise to a new fingertip sensor unseen during training. This is done by training a model on tactile images from four sensors and testing it on tactile images from a new sensor unseen during training (Tr4-TeU).

For each experiment, five models were trained to represent all distinct sensor data combinations. Evaluating all combinations highlights any underlying bias a model may have to learning features from specific tactile sensors.

Performance is gauged on the average mean absolute error (MAE) across all models trained using both TacViT and CNN approaches, an established metric in evaluating performance for these sensors \cite{lepora2020optimal, lloyd2024pose}. Additionally, the spread of MAE across all models is computed to evaluate the generalisation to a new unseen sensor.

\subsection{Data Collection}
% \begin{table}[b!]
% \centering
% \caption{Pose and parameter ranges used for training data collection.}
%     \begin{tabular}{|c|c|}
%     \hline
%     \textbf{Pose} & \textbf{Ranges} \\ \hline
%     $x$                & $[-2, 2]$\,mm       \\
%     $y$                & $[-2, 2]$\,mm       \\
%     $z$                & $[0,4]$\,mm         \\
%     $R_x$           & $[-20^\circ, 20^\circ]$ \\
%     $R_y$            & $[-20^\circ, 20^\circ]$ \\
%     $R_z$           & -                 \\ \hline
%     \end{tabular}
%     \label{tab:pose_ranges}
% \end{table}

The data collection was based on previous work with training convolutional neural networks with the TacTip family of sensors~\cite{lepora2020optimal}. Training data was collected by mounting each tactile sensor to the wrist of a Universal Robots UR5 industrial robot arm and contacting the sensor with a uniform flat surface at pre-determined poses. 3000 six degree-of-freedom Cartesian end-effector movements, which form the labels used in training, are generated within the following ranges: $x,y \in [-2, 2]$\,mm, $z \in [0,4]$\,mm, $R_x,R_y \in [-20^\circ, 20^\circ]$. The force ranges derive from these pose ranges and are approximately $F_x,F_y \in[-3, 3]\,$N and $F_z\in[0,10]\,$N. The robot is moved so that the sensor is brought into contact with the surface at the specified poses. The sensor is then sheared tangentially across the surface according to a displacement $(x,y)$, a practice that allows the model to ignore these perturbations, making it more robust to surface-dependent shear during servoing tasks \cite{lepora2020optimal}. Following a short pause, the resultant tactile image is captured and stored.

% \begin{enumerate}
%     \item Sensor is oriented at angles $R_x$ and $R_y$ about the $x$- and $y$-axis of the wrist frame above the surface.
%     \item The sensor is brought into contact with the surface at this orientation to a depth $z$.
%     \item  Introducing unlabeled shear perturbations into the dataset is standard practice when collecting TacTip data in this way, as it allows 
%     \item 
% \end{enumerate}

\subsection{Training}
TacViT and CNN models were trained using a PC running Windows 10 on an Nvidia 3070 GPU with CUDA 11.8. The software was developed in Python 3.11, using PyTorch 2.6 for learning.

Both models are trained using a supervised learning approach, where the model learns to predict depth ($z$), orientation ($R_x, R_y$), and contact force ($F_x, F_y, F_z$) from tactile images. In total, 15000 tactile images were used, collected from 5 sensors (3000 per sensor). Data are used in a 80/20 split for training/validation, where 20\% of the training data are held-out to prevent overfitting. The training process for the TacViT model consists of the following stages:

\subsubsection{Forward Pass}
Given an input tactile image $X$, it is first divided into non-overlapping patches of size $P \times P$. Each patch is linearly embedded into a fixed-dimensional feature space: $E = W_p X_p + b_p$, where $X_p$ represents a patch, $W_p$ is the learnable weight matrix, and $b_p$ is the bias term. These patch embeddings are then passed through the Vision Transformer encoder, consisting of multiple layers of self-attention and feedforward networks. The final output representation is extracted and passed to the regression head: $\hat{y} = f_\theta(E)$, where $\hat{y}$ represents the predicted depth and orientation, and $f_\theta$ denotes the transformer model with learnable parameters $\theta$.

\subsubsection{Fine-tuning with LoRA}

To improve computational efficiency, we apply Low-Rank Adaptation (LoRA) during fine-tuning \cite{hu2022lora, zeng2023expressive}. Instead of updating all transformer weights, LoRA introduces low-rank matrices into the self-attention layers: $\Delta W = W_A W_B$, where $W_A \in \mathbb{R}^{d \times r}$ and $W_B \in \mathbb{R}^{r \times d}$ are trainable low-rank matrices, with $r \ll d$. These matrices capture task-specific adaptations while keeping the pre-trained ViT parameters mostly frozen, allowing for more stable fine-tuning on tactile images.

Initially, only the regression head is trained while the transformer layers are kept frozen. After several epochs, selected transformer layers are unfrozen, enabling the model to fine-tune deeper representations. This gradual unfreezing strategy prevents catastrophic forgetting and stabilizes training.

\subsubsection{Loss Computation and Backpropagation}
The model is trained using Mean Squared Error (MSE) loss, calculating the difference between the predicted and ground truth values: $L = \frac{1}{N} \sum_{i=1}^{N} (\hat{y}_i - y_i)^2$, where $N$ is the batch size, $y_i$ is the ground truth label, and $\hat{y}_i$ is the model prediction.

To update model parameters, the gradients of the loss function with respect to each parameter are backpropagated through the transformer layers, adjusting the self-attention weights, feedforward layers, and the regression head. The parameters are updated using the Adam optimizer.

% \begin{equation}
% \theta \leftarrow \theta - \eta \frac{\partial L}{\partial \theta},
% \end{equation}

% where $\eta$ is the learning rate.

\subsubsection{Validation Process}
Experiments Tr1-Te1 and Tr5-Te1 use hold-out cross-validation where 20\% of the data across one or all tactile sensors is held back for testing. Experiment Tr4-TeU uses a form of 5-fold cross-validation, where the data from one of the five sensors is used to test and the other 4 sensors used for training (permuting the test sensor to find an average performance over all sensors).
% To ensure model robustness and prevent overfitting, TacViT is validated on a held-out dataset after each training epoch (20\% of training data). 
% The validation process consists of:
% \begin{itemize}
% \item \textbf{Forward pass on validation set}: The trained model is used to generate predictions $\hat{y}_v$ on unseen validation data $X_v$.
% \item \textbf{Validation loss computation}: The MSE loss on validation data is computed as:

% \begin{equation}
% L_v = \frac{1}{M} \sum_{j=1}^{M} (\hat{y}_{v,j} - y_{v,j})^2,
% \end{equation}

% where $M$ is the number of validation samples.
% \item \textbf{Early stopping and learning rate adjustment}: If validation loss does not improve for a set number of epochs, training is halted to prevent overfitting. Additionally, the learning rate is reduced based on validation loss trends:

% \begin{equation}
% \eta \leftarrow \eta \cdot \lambda,
% \end{equation}

% where $\lambda$ is a decay factor.

% \end{itemize}

The key hyperparameters and structural details of TacViT and CNN are outlined in Tables \ref{tab:hyperparams}, and \ref{tab:cnn_hyperparams} respectively. 

\begin{table}[h]
\centering
\caption{TacViT Model Hyperparameters and Architecture}
\label{tab:hyperparams}
\begin{tabular}{|l|l|}
\hline
\textbf{Parameter} & \textbf{Value} \\
\hline
Image Resolution & $224 \times 224$ \\
Patch Size & 16 $\times 16$ \\
Number of Patches & 196 \\
Embedding Dimension & 768 \\
Number of Transformer Layers & 12 \\
Number of Attention Heads & 12 \\
MLP Hidden Dimension & 3072 \\
Number of Regression Layers & 4 \\
Dropout coefficient & 0 \\
Activation Function & ReLU \\
Optimizer & Adam \\
Learning Rate & 1e-4 \\
Weight Decay & 1e-5 \\
Batch Size & 16 \\
Loss Function & Mean Squared Error (MSE) \\ \hline
\end{tabular}
\end{table}

% Alonsgide the ViT model, a comparative model trained using a simple CNN was generated. This model was trained using the same principles on the same data, i.e. 4 domains split 80/20 for training/validation and 1 domain reserved for testing, with labels encoded to a normalized range of $[-1,1]$. 

\begin{table}[h!]
\centering
\caption{CNN Hyperparameters}
    \begin{tabular}{|l|l|} 
    \hline
    \textbf{Parameter}                    & \textbf{Value}   \\ \hline
    Number of convolutional hidden layers & 4                \\
    Number of convolutional kernels       & 480              \\
    Number of fully connected layers      & 2                \\
    Number of dense hidden layer units    & 512              \\
    Hidden layer activation function      & ReLU             \\
    Dropout coefficient                   & 0                \\
    Batch size                            & 16               \\
    Adam decay                            & $1\times10^{-6}$ \\
    Adam $\beta_1$                        & 0.9              \\
    Adam $\beta_2$                         & 0.999            \\ \hline
    \end{tabular} \label{tab:cnn_hyperparams}
\end{table}

\section{Results}

\begin{table*}
\centering
\renewcommand{\arraystretch}{1.2} 
\setlength{\tabcolsep}{3pt}
\caption{MAE of predicted contact pose and force for all three evaluation experiments (Tr1-Te1, Tr5-Te1, Tr4-TeU) tested on individual five sensors of the robot hand using the TacCNN and TacViT models.}
\label{tab:cnn_tacvit_results}
\arrayrulecolor{black}
\begin{tabular}{|c|c|c|c|c|c|c|c|c|c|c|c|c|c|} 
\hline
\multicolumn{2}{|c|}{{\cellcolor[rgb]{0.396,0.396,0.396}}} & \multicolumn{6}{c|}{{\cellcolor[rgb]{0.608,0.608,0.608}}\textbf{CNN}} & \multicolumn{6}{c|}{{\cellcolor[rgb]{0.608,0.608,0.608}}\textbf{TacViT}} \\ 
\hline
\rowcolor[rgb]{0.937,0.937,0.937} \multicolumn{1}{|l}{{\cellcolor[rgb]{0.396,0.396,0.396}}} & {\cellcolor[rgb]{0.608,0.608,0.608}}\textbf{Sensor \#} & \textbf{$z \,(mm)$} & \textbf{$R_x (\degree)$} & \textbf{$R_y (\degree)$} & \textbf{$F_x \,(N)$} & \textbf{$F_y \,(N)$} & \textbf{$F_z \,(N)$} & \textbf{$z \,(mm)$} & \textbf{$R_x (\degree)$} & \textbf{$R_y (\degree)$} & \textbf{$F_x \,(N)$} & \textbf{$F_y \,(N)$} & \textbf{$F_z \,(N)$} \\ 
\hline
{\cellcolor[rgb]{0.753,0.753,0.753}} & \#1 & 0.1251 & 1.1292 & 1.0961 & 0.1699 & 0.1688 & 0.4296 & 0.0675 & 1.5379 & 1.5292 & 0.1943 & 0.1970 & 0.4829 \\ 
\cline{2-14}
{\cellcolor[rgb]{0.753,0.753,0.753}} & \#2 & 0.1013 & 0.8702 & 0.8642 & 0.1110 & 0.1093 & 0.2585 & 0.0599 & 0.7189 & 0.7407 & 0.0953 & 0.0725 & 0.1945 \\ 
\cline{2-14}
{\cellcolor[rgb]{0.753,0.753,0.753}} & \#3 & 0.0959 & 0.7548 & 0.7464 & 0.1118 & 0.1071 & 0.2722 & 0.0542 & 0.6676 & 0.6163 & 0.0937 & 0.0992 & 0.2319 \\ 
\cline{2-14}
{\cellcolor[rgb]{0.753,0.753,0.753}} & \#4 & 0.0860 & 0.7934 & 0.7964 & 0.1145 & 0.1137 & 0.2549 & 0.0735 & 0.9455 & 0.8948 & 0.0974 & 0.1093 & 0.2207 \\ 
\cline{2-14}
\multirow{-5}{*}{{\cellcolor[rgb]{0.753,0.753,0.753}}\textbf{Tr1-Te1}} & \#5 & 0.0981 & 0.7263 & 0.7482 & 0.1208 & 0.1225 & 0.3010 & 0.0678 & 1.2899 & 0.6957 & 0.0939 & 0.1012 & 0.2931 \\ 
\hline
{\cellcolor[rgb]{0.753,0.753,0.753}} & \#1 & 0.1111 & 1.0749 & 1.0635 & 0.2038 & 0.2063 & 0.5376 & 0.0637 & 3.6944 & 2.7510 & 0.1934 & 0.1725 & 0.5662 \\ 
\cline{2-14}
{\cellcolor[rgb]{0.753,0.753,0.753}} & \#2 & 0.0775 & 1.3379 & 1.4834 & 0.1279 & 0.1041 & 0.3422 & 0.0584 & 1.1625 & 0.8478 & 0.0822 & 0.0869 & 0.2104 \\ 
\cline{2-14}
{\cellcolor[rgb]{0.753,0.753,0.753}} & \#3 & 0.0630 & 0.9290 & 1.4435 & 0.0917 & 0.0961 & 0.2373 & 0.0634 & 1.0925 & 0.9822 & 0.0994 & 0.0778 & 0.2156 \\ 
\cline{2-14}
{\cellcolor[rgb]{0.753,0.753,0.753}} & \#4 & 0.0763 & 0.7292 & 0.7936 & 0.1207 & 0.1205 & 0.2747 & 0.0641 & 1.2412 & 1.1337 & 0.1047 & 0.0700 & 0.2333 \\ 
\cline{2-14}
\multirow{-5}{*}{{\cellcolor[rgb]{0.753,0.753,0.753}}\textbf{Tr5-Te1}} & \#5 & 0.0838 & 0.7499 & 0.7756 & 0.1184 & 0.1244 & 0.3453 & 0.0518 & 1.1398 & 1.2224 & 0.0962 & 0.0908 & 0.2810 \\ 
\hline
{\cellcolor[rgb]{0.753,0.753,0.753}} & \#1 & 0.2455 & 8.9341 & 6.0919 & 0.5508 & 0.6224 & 1.2887 & 0.0821 & 6.9181 & 6.3864 & 0.3467 & 0.2811 & 0.8304 \\ 
\cline{2-14}
{\cellcolor[rgb]{0.753,0.753,0.753}} & \#2 & 0.1767 & 4.0040 & 3.3047 & 0.2643 & 0.2992 & 0.4982 & 0.0873 & 1.6298 & 1.9585 & 0.1483 & 0.1327 & 0.4179 \\ 
\cline{2-14}
{\cellcolor[rgb]{0.753,0.753,0.753}} & \#3 & 0.1855 & 3.3189 & 3.1848 & 0.3123 & 0.3071 & 0.5289 & 0.0572 & 1.3993 & 1.4604 & 0.0941 & 0.0925 & 0.2463 \\ 
\cline{2-14}
{\cellcolor[rgb]{0.753,0.753,0.753}} & \#4 & 0.1510 & 3.9860 & 4.1689 & 0.3220 & 0.4329 & 0.4769 & 0.0703 & 1.9415 & 1.6366 & 0.1649 & 0.1616 & 0.3364 \\ 
\cline{2-14}
\multirow{-5}{*}{{\cellcolor[rgb]{0.753,0.753,0.753}}\textbf{Tr4-TeU}} & \#5 & 0.2991 & 8.5944 & 6.7381 & 0.5083 & 0.4673 & 1.3176 & 0.1009 & 1.6814 & 1.8652 & 0.1690 & 0.1568 & 0.3381 \\ 
\hline
\end{tabular}
\end{table*}

\begin{table*}[t]
\arrayrulecolor{black}
\renewcommand{\arraystretch}{1.2}
\setlength{\tabcolsep}{4pt}
\centering
\caption{Comparison of Mean MAE of predicted contact pose and force for CNN and TacViT models.}
\label{tab:results}
\begin{tabular}{|>{\centering\arraybackslash}p{2.8cm}|>{\centering\arraybackslash}p{1.5cm}|
>{\centering\arraybackslash}p{0.9cm}|>{\centering\arraybackslash}p{0.9cm}|>{\centering\arraybackslash}p{0.9cm}|
>{\centering\arraybackslash}p{0.9cm}|>{\centering\arraybackslash}p{0.9cm}|>{\centering\arraybackslash}p{0.9cm}|
>{\centering\arraybackslash}p{0.9cm}|>{\centering\arraybackslash}p{0.9cm}|>{\centering\arraybackslash}p{0.9cm}|
>{\centering\arraybackslash}p{0.9cm}|>{\centering\arraybackslash}p{0.9cm}|>{\centering\arraybackslash}p{0.9cm}|}
\hline
\rowcolor[HTML]{C0C0C0} 
\multicolumn{2}{|c|}{\cellcolor[HTML]{656565}} 
& \multicolumn{6}{c|}{\cellcolor[HTML]{C0C0C0}\textbf{CNN}} 
& \multicolumn{6}{c|}{\cellcolor[HTML]{C0C0C0}\textbf{TacViT}} \\ \cline{3-14} 
\rowcolor[HTML]{EFEFEF} 
\multicolumn{2}{|c|}{\multirow{-2}{*}{\cellcolor[HTML]{656565}}} 
& \textbf{$z\,(mm)$} & \textbf{$R_x\,(\degree)$} & \textbf{$R_y\,(\degree)$} 
& \textbf{$F_x\,(N)$} & \textbf{$F_y\,(N)$} & \textbf{$F_z\,(N)$} 
& \textbf{$z\,(mm)$} & \textbf{$R_x\,(\degree)$} & \textbf{$R_y\,(\degree)$} 
& \textbf{$F_x\,(N)$} & \textbf{$F_y\,(N)$} & \textbf{$F_z\,(N)$} \\ \hline
% ----------- Tr1-Te1 -----------
\multicolumn{1}{|c|}{\cellcolor[HTML]{C0C0C0}\textbf{Tr1-Te1}}
& \cellcolor[HTML]{EFEFEF}\textbf{Mean MAE}
& 0.1013 & 0.8548 & 0.8503 & 0.1256 & 0.1243 & 0.3832
& 0.0646 & 1.0320 & 0.8953 & 0.1149 & 0.1158 & 0.2846 \\ \hline
% ----------- Tr5-Te1 -----------
\multicolumn{1}{|c|}{\cellcolor[HTML]{C0C0C0}\textbf{Tr5-Te1}}
& \cellcolor[HTML]{EFEFEF}\textbf{Mean MAE}
& 0.0823 & 0.9642 & 1.1119 & 0.1325 & 0.1303 & 0.3474
& 0.0625 & 1.6514 & 1.3855 & 0.1207 & 0.1065 & 0.3344 \\ \hline
% ----------- Tr4-TeU -----------
\multicolumn{1}{|c|}{\cellcolor[HTML]{C0C0C0}\textbf{Tr4-TeU}}
& \cellcolor[HTML]{EFEFEF}\textbf{Mean MAE}
& 0.2116 & 5.7675 & 4.6977 & 0.3915 & 0.4258 & 0.8221
& 0.0796 & 2.7140 & 1.8330 & 0.1846 & 0.1649 & 0.4338 \\ \hline
\end{tabular}
\end{table*}

% ============================================

Results are shown for all three evaluation methods and for each of the five sensors. Table \ref{tab:cnn_tacvit_results} shows the mean average error (MAE) of all pose and force variable predictions for the CNN and TacViT models. Table \ref{tab:results} shows the overall mean MAE across all predicted variables for both CNN and TacVit models. Means are calculated from the distribution of MAE across the models trained for each experiment, which encompass all distinct combinations used in training.

\subsection{Tr1-Te1: Train and test on single sensor}

In the Tr1-Te1 experiments, where training and validation are performed on individual sensor data, performance is generally comparable with TacViT performing slightly worse overall. The performance of both approaches on this experiment is good, exhibiting low MAE (high accuracy) and low standard deviation (consistency across multiple domain combinations). Previous studies have shown that $z_{\rm MAE} < 0.1$\,mm, $(R_x,R_y)_{\rm MAE}<2.5\degree$ and $F_{\rm MAE} < 1$N is considered to be acceptable \cite{lepora2020optimal}.

\subsection{Tr5-Te1: Train on five sensors and test on one seen sensor}

In the Tr5-Te1 experiments, where the model is trained on data from five sensors and tested on data from a single seen sensor, performance is again generally comparable between the two methods but it is better than the Tr1-Te1 experiment.

\subsection{Tr4-TeU: Train on four sensors and test on one unseen sensor}

In the Tr4-TeU experiment (where models are tested on data from an unseen sensor) the CNN performance is noticeably worse, exhibiting mean MAEs an order of magnitude higher than seen at baseline, as well as distributions of MAE up to 2.2 standard deviations from the mean. Conversely, TacViT's performance worsens but to a much lesser extent. Mean MAEs are higher than at Tr5-Te1, but accuracy is still good overall within the expected error values for robotics applications. In addition, TacViT retains much lower standard deviations than the CNN approach, showing better consistency across domain combinations.

\begin{figure*}[]
        \centering
        \includegraphics[width = 0.99\textwidth]{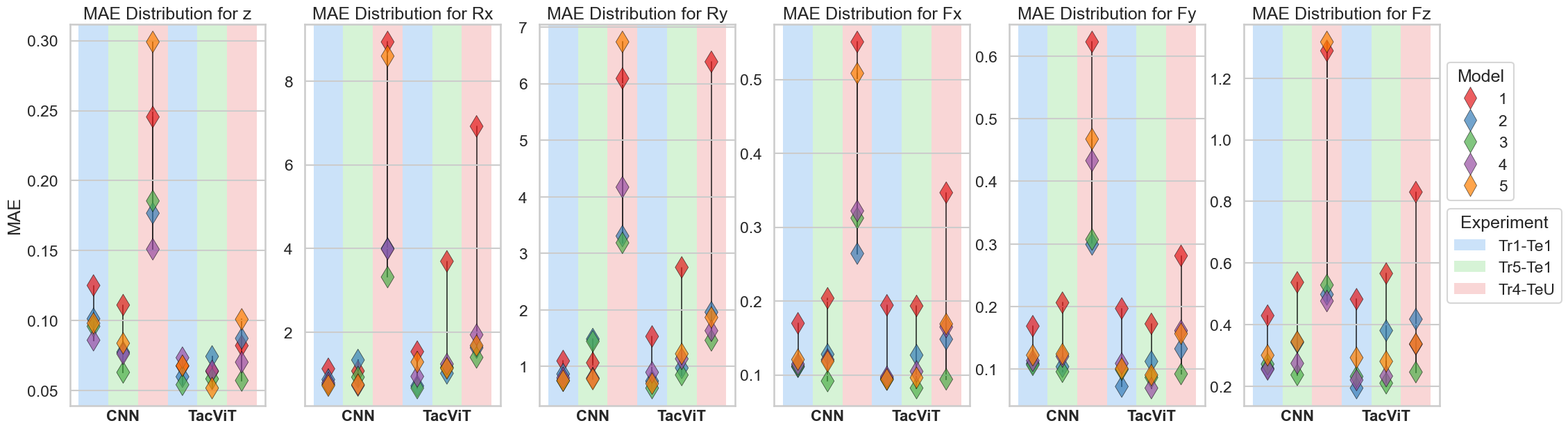}
        \caption{Strip plots showing the distribution of mean absolute error (MAE) using TacViT and CNN on the three evaluation experiments. Each diamond point represents each model trained per method and per experiment (encompassing all distinct training combinations).}
        \label{fig:result_2}
        \vspace{-1em}
\end{figure*}

The above results are further visualized in Fig. \ref{fig:result_2}, which shows strip plots of MAE for predicted pose and force variables for all models trained for all three experiments Tr1-Te1, Tr5-Te1 and Tr4-TeU. We see that in the Tr1-Te1 and Tr5-Te1 experiments, points for both CNN and TacViT are of low value and relatively tightly clustered. For the Tr4-TeU experiment, the MAEs for the CNN approach are much larger and exhibit a much larger spread. In contrast, points for TacViT are higher than its Tr5-Te1 (as expected) but only by a small amount and still remain clustered relative to the equivalent CNN results.

% \begin{figure}[h!]
%         \centering
%         \includegraphics[width = 0.48\textwidth]{images/result_2.jpg}
%         \vspace{-1em}
%         \caption{Strip plots showing the difference in performance between TacViT and a CNN on a baseline pose estimation task (tested on domains seen during training) and a domain adaptation (DA) pose estimation task (tested on domains unseen during training). Performance is measured using mean absolute error (MAE) with points shown for each model trained per method and experiment (encompassing all distinct training domain combinations).}
%         \label{fig:result_2}
%         \vspace{-1em}
% \end{figure}

\begin{figure*}[!t]
        \centering
        \includegraphics[width = 0.95\textwidth]{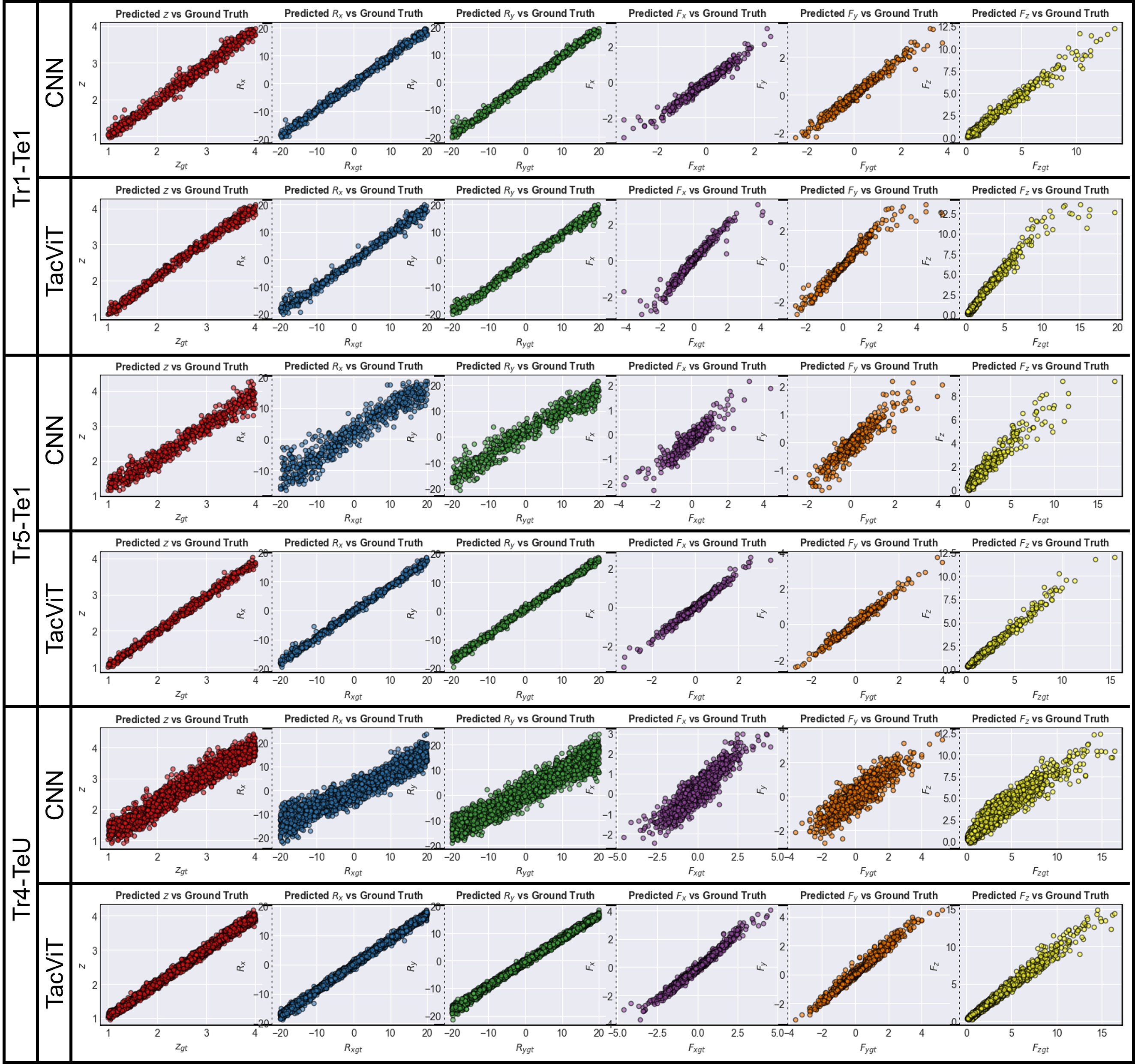}
        \caption{Example scatter plots displaying the predicted vs ground truth value for each pose and force parameter for each experiment using TacViT and CNN. Significant performance degradation can be seen in the CNN, whereas TacViT adapts better to data from an unseen domain.}
        \label{fig:result_1}
        \vspace{-1em}
\end{figure*}

Fig. \ref{fig:result_1} shows the difference in performance of each architecture in practice. The figure shows typical examples of predicted vs ground truth values for each experiment for both CNN and TacViT approaches. The plots for both methods in the Tr5-Te1 experiment are similar, as expected, with slightly more noise evident in TacViT. The performance degradation of the CNN approach on the Tr4-TeU is clear, with predictions becoming significantly noisier, whilst also failing to predict values in the correct range for $z$ (prediction value starts at 2mm rather than 1mm). Conversely, we see a relatively small increase in noise for TacViT and prediction ranges correct, indicating significantly greater robustness in generalising to an unseen sensor than using a CNN.

\section{Discussion and Conclusions}
% \subsection{Generalization Benefits of ViTs}
Our results confirm that TacViT enables generalization to new, unseen sensors - an essential capability for scalable tactile sensing in multi-fingered robot hands. This removes the need for retraining models for each individual fingertip sensor, significantly reducing deployment time and data collection effort. We have shown that ViTs generalize significantly better to new unseen sensors than CNNs by evaluating performance across multiple TacTip sensors mounted on a five-fingered robot hand. We attribute the superior performance of ViTs over CNNs to two key factors. {Global attention mechanisms} allow ViTs to better integrate information across disparate regions of an image. In the case of VBTS, this facilitates improved generalization across different sensors when faced with sensor-specific variances in tactile images such as artifacts, glare or brightness differences (Fig. \ref{fig:tactip}). {Hierarchical feature learning} for VBTS makes models more robust to noise or artifacts introduced into tactile images by individual sensor variances. Studies have shown that this capability leads to robustness against domain-specific variances and allows for more effective adaptation to new, unseen domains \cite{zhou2021convnets, li2021benchmarking}.

Some issues include that ViTs typically require larger datasets to fully exploit their self-attention mechanisms \cite{papa2024survey}. Unlike CNNs, which can effectively learn hierarchical features from relatively limited data due to their strong inductive biases, ViTs rely on large-scale pre-training to achieve competitive performance\cite{dosovitskiy2020image}.  In this study, we were able to get reasonable results using a ViT model pretrained on a publicly available dataset of general images (ImageNet), fine tuned on tactile images. This is possible due to the global nature of ViT learning, making these large pre-trained models excellent at extracting features from unseen domains. Performance is likely to be improved if a ViT were trained from scratch on a large dataset of many thousands of tactile images. Whilst resource intensive, the results of this paper suggest this to be a worthwhile avenue for future work. Another future direction would be to tailor the architecture design to tactile-specific priors such as symmetry or surface topology, which are particularly relevant in multi-fingered hands with repetitive sensor geometries. While this study focuses on TacTip sensors, which are widely used and representative of vision-based tactile sensing, we acknowledge that evaluating TacViT on other vision-based tactile sensor types would be valuable to assess its broader applicability.

Overall, in this paper, we introduced TacViT: a vision transformer-based architecture designed for generalising to new unseen sensor data in pose and force prediction for TacTip-style vision-based tactile sensors. We demonstrated that TacViT enables robust inference on new, unseen sensors without retraining, addressing a key challenge in deploying vision-based tactile sensors at scale. This work represents a step toward practical, plug-and-play tactile sensing in robotic manipulation, particularly in multi-fingered hands where sensor interchangeability is critical. Our work compares to traditional convolutional neural networks by comparing baseline performance of both approaches when tested on data from sensors seen during training vs data from an unseen sensor, evaluating mean absolute error and standard deviation as performance metrics. These results highlight the potential benefit of using ViTs in VBTS applications where even minor variations between sensors can degrade CNN model performance. Whilst TacViT outperformed the CNN in this case, there is still room for improvement. 
Quickly deploying new vision-based tactile sensors in multi-fingered robot hands is a significant limiting factor in the field; this work represents a step towards solving that problem and closing the gap on mainstream adoption of high-resolution tactile sensing in robotic manipulation.

\bibliographystyle{IEEEtran}
\bibliography{ref}

\end{document}